\documentclass[letterpaper, 10 pt, conference]{ieeeconf}  

\IEEEoverridecommandlockouts                              

\usepackage{times}
\usepackage{dsfont}
\usepackage{multicol}
\usepackage[bookmarks=true]{hyperref}
\hypersetup{
    colorlinks=true,
    linkcolor=blue,
    filecolor=magenta,      
    urlcolor=black, 
    pdftitle={Overleaf Example},
    pdfpagemode=FullScreen,
    }
\usepackage[utf8]{inputenc} 
\usepackage[T1]{fontenc}    
\usepackage{hyperref}       
\usepackage{url}            
\usepackage{booktabs}       
\usepackage{amsfonts,bm}       
\usepackage{amssymb}
\usepackage{nicefrac}       
\usepackage{xcolor}         
\usepackage{bbm}

\usepackage{caption}
\usepackage{subcaption}

\usepackage{bbm}
\usepackage{lipsum}
\usepackage{algorithm}
\usepackage{algpseudocode}
\usepackage{graphicx}
\usepackage{framed}

\usepackage{siunitx}

\usepackage{authblk}
\usepackage[T1]{fontenc}
\usepackage[utf8]{inputenc}

\let\labelindent\relax
\usepackage{enumitem}

\usepackage{mathtools}
\graphicspath{ {images/} }

\usepackage{multicol}
\usepackage{svg}
\usepackage{multirow}

\title{\LARGE \bf
Graph-based 3D Collision-distance Estimation Network \\
with Probabilistic Graph Rewiring
}

\author{Minjae Song$^1$, Yeseung Kim$^1$, Min Jun Kim$^2$, and Daehyung Park$^1$\textsuperscript{\textdagger}
\thanks{$^1$M. Song, $^1$Y. Kim and $^1$D. Park are with the School of Computing, and $^2$M. J. Kim is with the School of Electrical Engineering, Korea Advanced Institute of Science and Technology, Korea ({\tt\small \{smj0398, yeseung.kim.88, minjun.kim, daehyung\}@kaist.ac.kr}). 
{\textsuperscript{\textdagger}}D. Park is the corresponding author. 
\\
This work was partly supported by Institute of Information \& communications Technology Planning \& Evaluation (IITP) grant funded by the Korea government (MSIT) (No.2022-0-00311, Development of Goal-Oriented Reinforcement Learning Techniques for Contact-Rich Robotic Manipulation of Everyday Objects), the KAIST Convergence Research Institute Operation Program, and the National Research Foundation of Korea (NRF) grants funded by the Korea government (MSIT) (No. 2021R1A4A3032834).}
}

\begin{document}

\maketitle
\thispagestyle{empty}
\pagestyle{empty}

\begin{abstract}
We aim to solve the problem of data-driven collision-distance estimation given 3-dimensional (3D) geometries. Conventional algorithms suffer from low accuracy due to their reliance on limited representations, such as point clouds. In contrast, our previous graph-based model, GraphDistNet, achieves high accuracy using edge information but incurs higher message-passing costs with growing graph size, limiting its applicability to 3D geometries. To overcome these challenges, we propose GDN-R, a novel 3D graph-based estimation network. GDN-R employs a layer-wise probabilistic graph-rewiring algorithm leveraging the differentiable Gumbel-top-$K$ relaxation. Our method accurately infers minimum distances through iterative graph rewiring and updating relevant embeddings. The probabilistic rewiring enables fast and robust embedding with respect to unforeseen categories of geometries. Through $41,412$ random benchmark tasks with $150$ pairs of 3D objects, we show GDN-R outperforms state-of-the-art baseline methods in terms of accuracy and generalizability. We also show that the proposed rewiring improves the update performance reducing the size of the estimation model. We finally show its batch prediction and auto-differentiation capabilities for trajectory optimization in both simulated and real-world scenarios.
\end{abstract}
\IEEEpeerreviewmaketitle

\section{Introduction}

The ability to detect collisions is the fundamental requirement for the safe and efficient operation of robotic systems in real-world scenarios. Such operation requires computing the minimum distance between geometries~\cite{vorndamme2017collision} (see Fig.~\ref{fig:main}). Traditional iterative collision search algorithms, such as Gilbert–Johnson–Keerthi (GJK)~\cite{gilbert1988fast}, have been widely adopted for finding distances between convex geometries. However, these iterative approaches encounter significant time and space complexities when dealing with complex geometries.

In contrast, data-driven distance estimation methods have gained attention due to their parallelization and auto-differentiation capabilities that can be facilitated by various learning and optimization applications. Typical approaches take point clouds of geometries as input and then output the minimum distance~\cite{chase2020neural, kim2023pairwisenet}. To account for swept volumes, approaches often take joint configuration~\cite{zhi2022diffco} or transformation~\cite{kim2023pairwisenet} of geometries as input data. However, using simple point-based representations has limitations when estimating accurate distances for non-convex geometries and often suffers from generalizability to new environments. Alternatively, recent studies have explored signed distance fields (SDFs)~\cite{koptev2022neural, michaux2023reachability}, offering continuous distance measures for query points. While SDFs provide continuous and smooth distance information, they often lack generalizability to new geometries and require computationally expensive training processes.

\begin{figure}[t]
 \centering
  \includegraphics[width=0.45\textwidth]{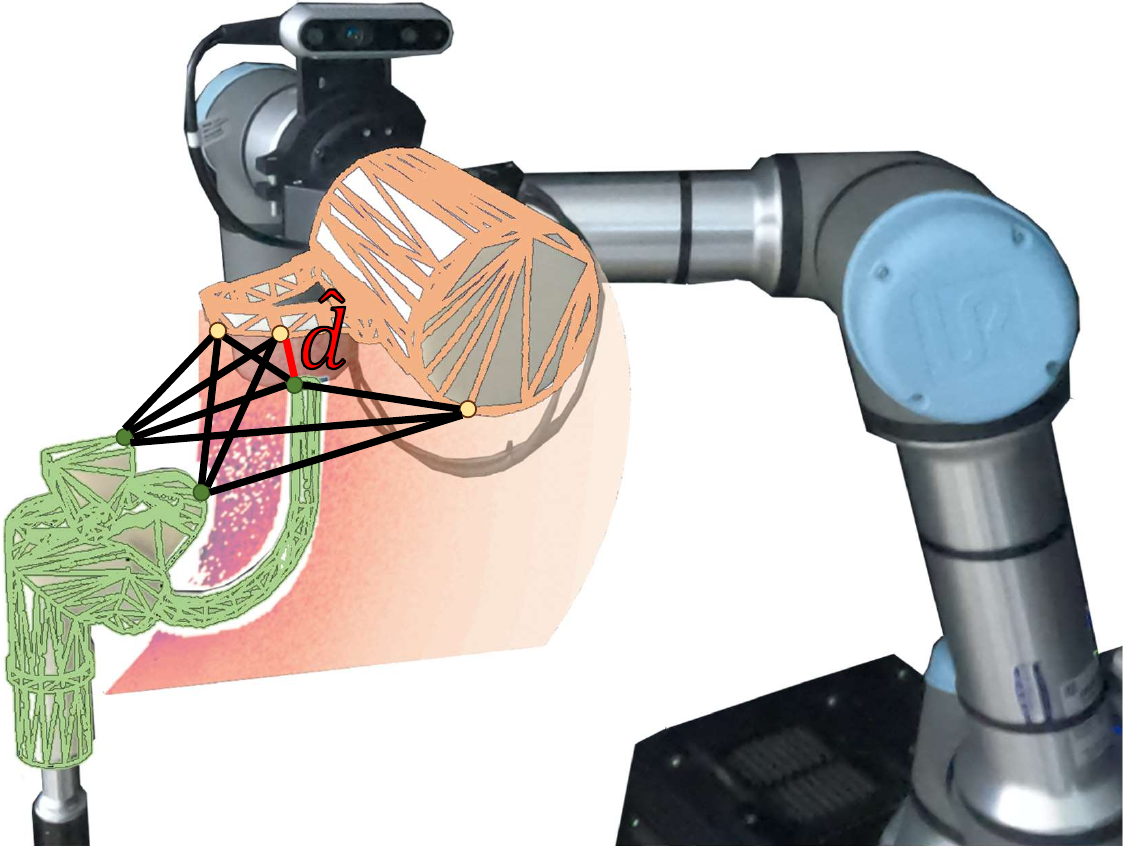}  \caption{A capture of the object-hanging experiment. Our graph-based collision-distance estimator, GDN-R, probabilistically rewires two graph geometries and precisely regresses the minimum distance and collision gradients between them.}
  \vspace{-2em}
  \label{fig:main}
\end{figure}

Instead, GraphDistNet~\cite{kim2022graphdistnet}, a data-driven estimator, regresses the minimum distance between two graph geometries by employing message-passing with distance-relevant embeddings. To pass information through wired graphs (i.e., a connected graph), GraphDistNet creates large bipartite structures with attention mechanisms. While the connections enable precise and scalable predictions, the wiring process becomes computationally intensive with larger graphs, limiting its application to 2-dimensional (2D) environments.

To resolve it, we introduce GDN-R, a novel graph-based distance estimation network with a layer-wise probabilistic rewiring method using the Gumbel top-$k$ relaxation~\cite{kool2019stochastic, xie2019subsets}. This method takes two graphed objects as input and then iteratively updates distance-related embeddings through a message-passing neural network (MPNN)~\cite{gilmer2020message}. The proposed rewiring method identifies the top $k$-largest probability of interconnection nodes to create higher connectivity of the bipartite structure, which involves inferring accurate minimum distance. Its differentiable nature also allows for the gradual update of the interconnection nodes using gradient descent. Moreover, the probabilistic sampling with the Gumbel noise allows the creation of random inter edges to explore unknown message-pass for information, robust to unforeseen geometries and overfitting to known geometries.   

We conduct extensive statistical evaluations by creating a collision-distance benchmark dataset based on the 3D object-hanging manipulation benchmark~\cite{you2021omnihang}. Our evaluation shows GDN-R accurately regresses minimum collision distances between object-hook geometries, while state-of-the-art data-driven baseline methods have low accuracy and difficulty in handling unseen categories of geometries. Further, we show that the graph-based distance regression can be extended to 3D environments by introducing an efficient model. We finally demonstrate the applicability of the learned distance estimation model and its gradient estimation for trajectory optimization. 

Our main contributions are as follows:
\begin{itemize}[leftmargin=*]
    \item We introduce a highly accurate 3D collision-distance estimator, GDN-R, leveraging graph representations. 
    \item We present a novel probabilistic rewiring algorithm using the Gumbel top-$k$ relaxation, which enables fast distance estimation robust to unforeseen categories of 3D geometries.
    \item We conduct various benchmark evaluations of our method against four state-of-the-art estimators using 3D geometries. We also demonstrate the application of GDN-R in trajectory optimization with manipulation scenarios.    
\end{itemize}

\section{Related Work}
\label{sec:related}
Traditional collision-distance methods use multiple levels of the filtering processes: broad and narrow phases. The broad phase involves coarse filtering, which excludes large collision-free primitives (i.e., cells~\cite{luque2005broad} or bounding volumes~\cite{quinlan1994efficient, larsson2006dynamic}). Then, the narrow phase of algorithms, such as GJK~\cite{gilbert1988fast}, precisely checks the intersection within the collision set. 
However, these algorithms often suffer from large meshes and collision queries due to their iterative search in a single thread. To mitigate the computational complexity, approaches often simplify target geometries into simpler primitives such as cuboids~\cite{benallegue2009fast}, spheres~\cite{schulman2014motion}, and voxels~\cite{lawlor2002voxel}, albeit at the cost of accuracy.

Alternatively, data-driven estimation algorithms have gained attraction for their efficiency in batch prediction of distances and their gradients. These algorithms often learn to map configurations to the minimum distance between a robot and obstacles~\cite{das2017fastron, zhi2022diffco, chase2020neural, koptev2022neural, munoz2023collisiongp, das2020forward}. A representative model is ClearanceNet~\cite{chase2020neural}, a neural network-based approach that enables parallelized and GPU-efficient batch computation. However, configuration mapping is challenging to generalize to scenarios involving moving or shape-changing obstacles. Recent works have shifted towards taking 3D point clouds of both the robot and obstacles as input~\cite{son2023local, kim2023pairwisenet, you2021omnihang, danielczuk2021object}. Although point-based representations exhibit low accuracy, particularly with complex 3D geometries, our method represents geometries as 3D graphs, leveraging edge information for precise distance estimation, similar to our previous work, GraphDistNet~\cite{kim2022graphdistnet}. 

On the other hand, researchers have introduced binary collision-detection algorithms~\cite{danielczuk2021object, son2023local}. The detectors do not provide neither distance nor gradients useful for optimization-based applications. Recent studies have demonstrated the potential of learning SDFs, offering continuous and smooth collision distances with query efficiency~\cite{koptev2022neural, danielczuk2021object, michaux2023reachability}. While SDF learning is challenging, distance estimation requires identifying collision points separately or using sphere-like point objects. However, our GDN-R directly computes the minimum distance between two geometries without identifying the exact collision point.


Lastly, our primary application lies in trajectory optimization (TO), a process aimed at synthesizing optimal trajectories satisfying specific constraints (e.g., collision avoidance)~\cite{yoon2023learning}. Most TO methods leverage gradient-descent techniques to expedite the search~\cite{zucker2013chomp, schulman2014motion}. For instance, CHOMP~\cite{zucker2013chomp} uses covariant gradient techniques precomputing the collision gradient of simple geometries through the robot's configuration space. In similar, TrajOpt~\cite{schulman2014motion} uses sequential convex optimization to calculate numerical gradients between the robot and environmental objects. However, conventional gradient computations are often challenging or unstable when handling complex geometries. Recent data-driven distance estimators, including GraphDistNet, offer approximated yet stable gradients, enabling efficient batch predictions. In this work, we show how our proposed method enhances the optimality and efficiency of the TO process.

\section{Background}
\label{sec:background}
\newcommand{\vx}{{\bf x}}
We briefly review how we embed necessary features through graph neural network (GNN) with 1-hop message passing. Let $G=(\mathcal{V},\mathcal{E})$ denote a graph composed of nodes $\mathcal{V}$ and edges $\mathcal{E}$. We establish an edge $e_{ij}\in \mathcal{E}$ when a node $v_i \in \mathcal{V}$ is adjacent to a node $v_j \in \mathcal{V}$. 
Here, the subscripts, $i$ and $j$, represent node indices. 
In addition, let $\vx_i \in \mathbb{R}^{d_v}$ denote the feature of node $v_i$, and $\vx_{ij} \in \mathbb{R}^{d_e}$ represent the feature vector of edge $e_{ij}$, where $d_v$ and $d_e$ denote user-defined feature dimensions.
Then, by aggregating features from the neighborhood, GNN updates node and edge features across layers, while also predicting the desired graph properties.  

\begin{figure*}[t]
 \center
  \includegraphics[width=\textwidth]{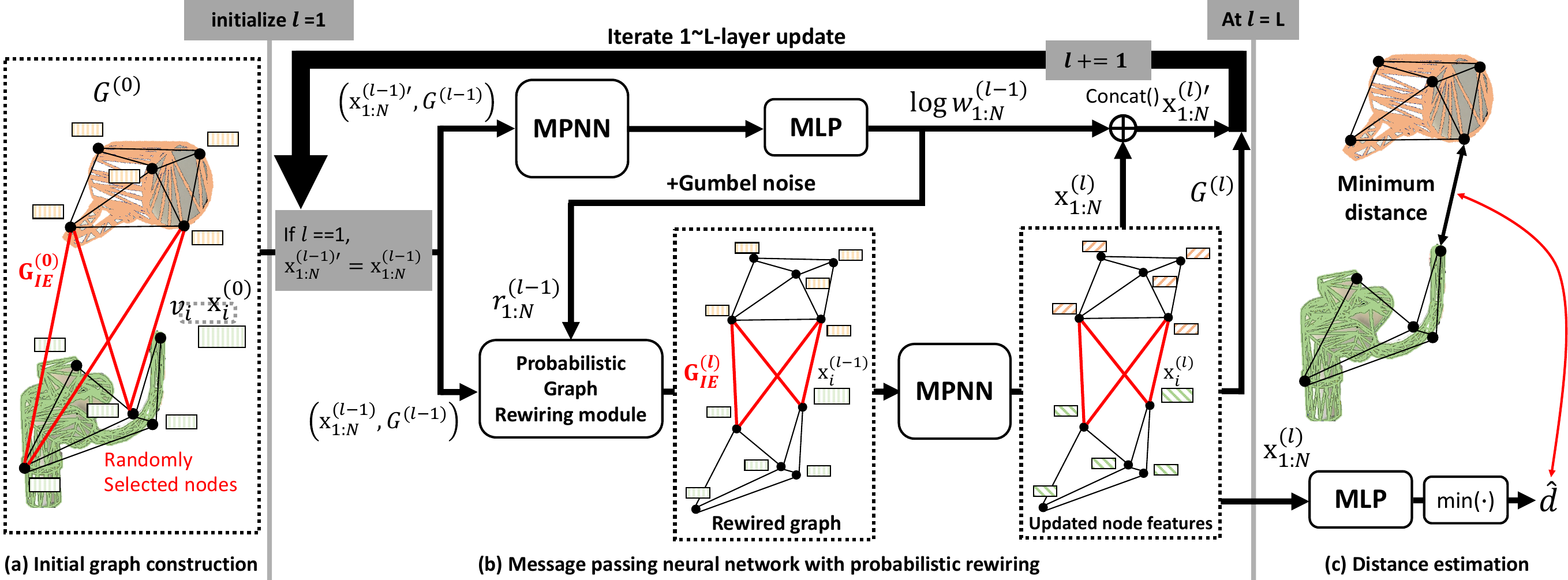}
  \caption{ Overall architecture of a graph-based collision-distance estimation network with a layer-wise probabilistic rewiring method (GDN-R). Given two graphed geometries, our GDN-R constructs a connected graph and iteratively updates distance-related embeddings through a message-passing neural network (MPNN). Alter $L$ times of passing operations, we estimate the minimum distance $\hat{d}$ using a readout MLP. Note that $\mathbf{x}_{1:N}^{(l)}=[\mathbf{x}_1^{(l)}, ... , \mathbf{x}_{N}^{(l)} ]$, $r_{1:N}^{(l-1)}= [ r_1^{(l-1)}, ... ,r_{N}^{(l-1)} ]$, and $\log w_{1:N}^{(l-1)}=[ \log w_{1}^{(l-1)}, ... , \log w_{N}^{(l-1)} ]$, where $N$ is the number of nodes in $G$.
  }
  \label{fig:overview}
\end{figure*}

In this work, we employ the message-passing neural network (MPNN)~\cite{gilmer2020message} to update features over layers. To update a node feature $\mathbf{x}_i^{(l+1)}$ at the $(l+1)$-th layer, MPNN uses three steps: \textit{message creation}, \textit{message aggregation}, and \textit{feature update}. The \textit{message creation} step builds neighbor-wise messages for the target node $v_i$ using multi-layer perceptron (MLP), $\phi$, in which the message from a neighbor node $v_j$ is $\mathbf{m}_{ij}=\phi (\mathbf{x}_i^{(l)}, \mathbf{x}_j^{(l)}, \mathbf{x}_{ij}^{(l)})$. Then, defining an aggregation operation $\square$ (e.g., $\sum$ or $\max$), the network aggregates all messages and updates the target node feature $\mathbf{x}_i^{(l+1)}$ via another MLP, $\gamma$. We represent the mathematical formulation of the node update process based on message passing~\cite{fey2019fast} as follows:
\begin{equation}
\mathbf{x}_i^{(l+1)}=\gamma \left( \mathbf{x}_i^{(l)}, \underset{j:e_{ij}\in \mathcal{E}}{\square}\phi\left(\mathbf{x}_i^{(l)},\mathbf{x}_j^{(l)},\vx_{ij}^{(l)}\right)\right).
\label{eq_mpnn}
\end{equation}
\section{Method: GDN-R}
\label{sec:method}

Consider the problem of determining the minimum collision distance $d$ between two triangle mesh objects, $o_1$ and $o_2$ as shown in Fig.~\ref{fig:overview}. Our proposed method, GDN-R, takes two graphed objects $(G_{o_1}, G_{o_2})$ and then iteratively finds distance-related embeddings (i.e., node and edge features) via MPNN using a novel inter-edge rewiring method. 
In the final \textit{read-out} layer, our method regresses the collision distance $\hat{d}$ and, if required, gradients. 

\subsection{Initial graph construction} \label{ssec:graphconstruction}
We create a connected graph $G$ for message passing by converting two input triangle meshes, $o_1$ and $o_2$, into graph representations: $G_{o_1}=(\mathcal{V}_{o_1},\mathcal{E}_{o_1})$ and $G_{o_2}=(\mathcal{V}_{o_2},\mathcal{E}_{o_2})$. In this work, we assume the independence of two graphs, i.e., $\mathcal{V}_{o_1}\cap \mathcal{V}_{o_2}=\emptyset$. We then initialize node features as tuples, 
\begin{align}
\vx_{i}^{(0)}&=(\mathbf{p}_{i}, ^{o_2}\mathbf{p}_{i},c_{o_1}) \quad \text{where }v_i \in \mathcal{V}_{o_1}  \nonumber \\
\vx_{j}^{(0)}&=(\mathbf{p}_{j}, ^{o_1}\mathbf{p}_{j},c_{o_2}) \quad \text{where }v_j \in \mathcal{V}_{o_2},  \nonumber 
\end{align}
where $\mathbf{p}_{i}\in\mathbb{R}^3$ is the $i$-th node coordinate in the global frame, $^{o_2}\mathbf{p}_{i}\in\mathbb{R}^3$ is the $i$-th node coordinate in the center frame of another object $o_2$, and $c_{o_1}\in\{0,1\}$ is a class value of the object $o_1$. We also calculate each edge feature by computing the difference between the source node feature and the target node feature: $\vx_{ij}^{(0)}=\vx_{i}^{(0)}-\vx_{j}^{(0)}$. This subtraction of node coordinates, $\mathbf{p}_{i}-\mathbf{p}_{j}$, yields relative coordinates that closely relate to the Euclidean distance between the two nodes. In addition, relative positions, such as $^{o_2}\mathbf{P}_i$ and $^{o_1}\mathbf{P}_j$, provide a preliminary estimate of the minimum distance between the selected node and the other object for selecting the subsequent interconnection nodes. The class value $c$ serves to distinguish the significance of message passing across objects.

To build a connected graph, we create a complete bipartite graph denoted as $G_{\text{IE}}=(\mathcal{V}_{\text{IE}}, \mathcal{E}_{\text{IE}})$, where each edge connects a node in $G_{o_1}$ to a node in $G_{o_2}$ or vice versa. Let $n_{\text{IE}}$ represent the count of interconnection nodes ($=|\mathcal{V}_{\text{IE}}|$), evenly distributed between $G_{o_1}$ and $G_{o_2}$. The edge set $\mathcal{E}_{\text{IE}}$ serves as passages for transferring relevant minimum-distance relevant geometric and semantic information between two graphs. We define the set as follows:
\begin{align}
    \mathcal{E}_{\text{IE}}=\{ e_{ij} |  (v_i \in \mathcal{V}_{o_1} \And v_j \in \mathcal{V}_{o_2}) \nonumber \\
    || (v_i \in \mathcal{V}_{o_2} \And v_j \in \mathcal{V}_{o_1})    \}.
\end{align}
In this initialization, we randomly sample $n_{\text{IE}}/2$ number of nodes from each graph, employing either uniform random sampling or informed priors. Furthermore, we establish edge features through feature subtraction to ensure structural consistency. Finally, we represent the connected graph as $G=G_{o_1}\cup G_{o_2} \cup G_{\text{IE}}$.

\subsection{Message passing with probabilistic rewiring} \label{ssec:message_passing}
GDN-R iteratively updates node features in the connected graph $G$ by probabilistically rewiring two input graphs, $G_{o_1}$ and $G_{o_2}$. Our MPNN is a modified version of Eq.~(\ref{eq_mpnn}) to aggregate the 1-hop neighborhood information of a target node $v_i$ and update its feature $\vx_i$ at the subsequent layer. The core idea of this process is that all that is needed to find the minimum distance between two objects is the minimum distance part of each object and the relative geometric information between them, which can be obtained by message-passing features from local parts. Consequently, we formulate the new MPNN at the $l$-th layer as follows:  
%
\begin{align}
    \vx_i^{(l+1)} &=  \sum_{j\in \mathcal{N}_{G} \left( i \right) } h_{\theta_M}^{(l)} \left( \vx_{i}^{\left(l \right)}, \vx_{j}^{\left(l \right)}, \vx_{ij}^{\left(l \right)} \right),
\label{node_embedding_update}
\end{align}

\begin{figure}[t]
 \centering
  \includegraphics[width=\linewidth]{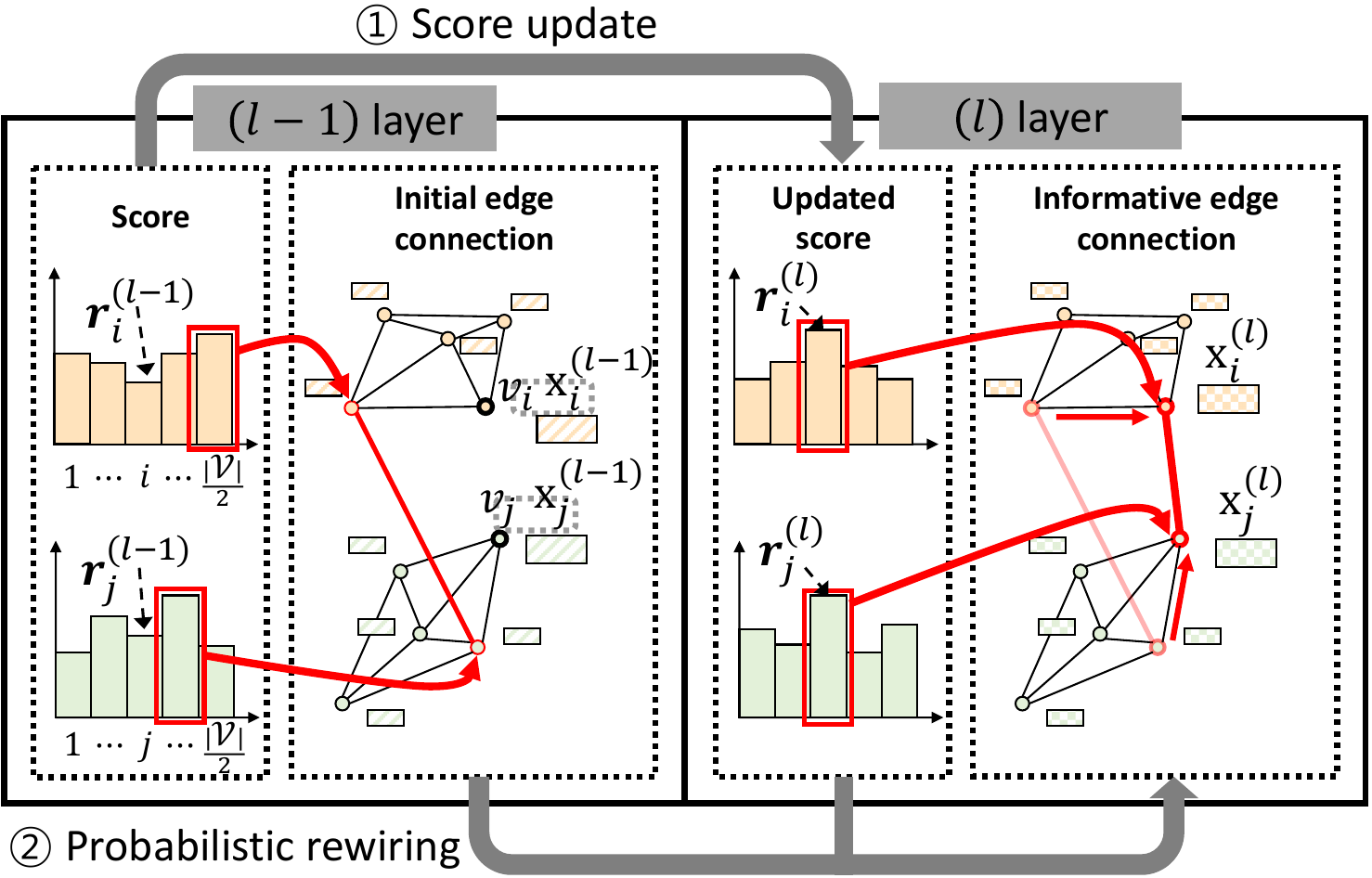}
  \caption{Exemplar illustration of probabilistic rewiring that gradually seeks nodes that hold essential geometric information to regress minimum distance. At each layer, our method updates the node-wise score utilizing the updated node feature. In the above example, though $i,j$-th node is not connected with inter edge, through message passing-based score update, the entire graph is updated toward informative connection.
  }
  \label{fig:main}
\end{figure}
%
where $\mathcal{N}_{G}(i)$ is the $1$-hop neighborhood node set of the node $v_i$ in the connected graph $G$, and $h_{\theta_M}^{(l)}$ is an MLP-based message creation function, parameterized by $\theta_M$. In this work, we use an MLP with dimensions $(16,16)$ and ReLU activation functions, and the message is passed twice, resulting in $\vx_i^{(l)}\in \mathbb{R}^{16}$ when $l \geq 1$.

To effectively update node features between the two graphs, we introduce a probabilistic rewiring method leveraging a relaxed version of the Gumbel top-$k$ trick~\cite{maddison2014sampling, maddison2016concrete, xie2019subsets}. In each object graph at the $l$-th layer, this method draws an ordered sample of size $n_{\text{IE}}/2$ without replacement from a perturbed log-probability distribution to construct a new $G_{\text{IE}}^{(l)}$. We denote this distribution, a perturbed score, as $r_i^{(l)} = \log  w_i^{(l)}  - \log\left( -\log u_i \right)$ where $w_i^{(l)}$ and $u_i$ are a non-negative scalar weight (i.e., probability) and a Gumbel noise $u_i\sim Gumbel(0,1)$, respectively. A node score $r_i^{(l)}$ is indicative of the relevance of its feature information for estimating the minimum distance, with higher scores signifying greater importance. To update weights across layers, we introduce weight initialization and update schemes as follows: 
%
\begin{itemize}[leftmargin=*]
\item \textbf{Initialization ($l=1$)}: We compute initial unperturbed scores $\log w_i^{(l)}$ by embedding node features of the initial graph $G^{(l)}$ through a single layer MPNN following Eq.~(\ref{node_embedding_update}). This process updates and compresses node features, resulting in $[\vx_1^{(l)'}, ... , \vx_{|\mathcal{V}|}^{(l)'}]$. We then map each feature to a score using $\log w_i^{(l)} = M_R^T \vx_i^{(l)'}$. For this, we use a $(4,4)$ MLP with ReLU activation functions in the MPNN, and $M_R$ represents a trainable projection matrix $(\in \mathbb{R}^{4\times 1})$. 
\item \textbf{Update ($l>1$)}: In subsequent iterations, we iteratively update unperturbed scores, $\log w_i^{(l)}$, by embedding the concatenated vector of the current node feature and the previous score, $[\vx_i^{(l)}, \log w_i^{(l-1)}]\in\mathbb{R}^{17}$,  through another single layer MPNN obtaining an updated node feature $\vx_i^{(l)'}$.  Then, as the initialization, we map each feature to a score using $\log w_i^{(l)} = M_R^T \vx_i^{(l)'}$. For this, we use a $(17,4)$ MLP with ReLU for the MPNN. 
\end{itemize}
Then, to sample interconnection nodes for $G_{\text{IE}}^{(l)}$, we use a top-$k$ relaxation method~\cite{xie2019subsets} that is differentiable with respect to the score $r_i^{(l)}$. This method allows using a temperature parameter $\tau$ where the randomness of the sampling procedure increases with higher $\tau$. In this work, we either fix or gradually decrease $\tau$. Across layers, our rewiring method improves the inter-edge bipartite graph $G_{\text{IE}}^{(l)}$ by re-sampling interconnection points on each graph, resulting in improved feature inference. This helps reduce the number of feature updates while preventing over-smoothing of features. In addition, probabilistic rewiring involves connecting high-score nodes between two graphs and maintaining the exploration of these connections by probabilistically connecting relevant random nodes.






\subsection{Training and test of distance estimation} \label{ssec:dist_est}
Following the last message-passing operation, we estimate the minimum-distance value $d$ using the readout network, as depicted in Fig.~\ref{fig:overview}. The network converts each node feature $\vx_i^{(L)} \in \mathbb{R}^{\times 16}$ from the final $L$-th layer into a node's distance $d_i$ from the target object, using a trainable projection matrix $M_{out} \in \mathbb{R}^{16\times 1}$. We then select the minimum value as $d=\min [d_1, ... d_{|\mathcal{V}|}] = \min M_{out}^T [\vx_1^{(L)}, ..., \vx_{|\mathcal{V}|}^{(L))}]$. 

For training, we optimize the parameters of GDN-R using two loss functions: $L_{\text{MSE}}$ and $L_{\text{WIRE}}$. $L_{\text{MSE}}$ is the mean-squared error loss between the predicted distance $d_i$ and the ground-truth distance $d_{true}$ which is the minimum distance between two objects. $L_{WIRE}$ quantifies the optimal connectivity of the wiring process. It calculates the summation of layer- and node-wise weighted regression errors,
\begin{align}
L_\text{WIRE}= \sum_{l=1}^{L}\sum_{i=1}^{|\mathcal{V}|} |d_{true}- d_i| \cdot r^{(l)}_i,
\label{eq:j_rd}
\end{align} 
where $L$ is the number of layers in GDN-R. For efficient backpropagation, we alternate $L_{\text{MSE}}$- and $L_{\text{WIRE}}$-based backpropagation during training.

\label{gnn}
\section{Experimental Setup}
\label{sec:eval_setup}



\subsection{Quantitative studies}

We use the OmniHang benchmark to evaluate the performance of distance estimation across various object-hanging scenarios, as illustrated in Fig.~\ref{fig:main}. The benchmark provides meshes for objects and hooks with predefined hanging goal poses. Our evaluation encompasses five object categories: \textit{cap}, \textit{headphone}, \textit{bag}, \textit{scissor}, and \textit{mug}, each containing ten different object geometries. We also use three U-shaped hooks to hang the selected objects. We perform distance evaluation by dividing the $150$ object-hook pairs following the seen and unseen categories.

\begin{figure}[t]
 \center
  \includegraphics[width=\linewidth]{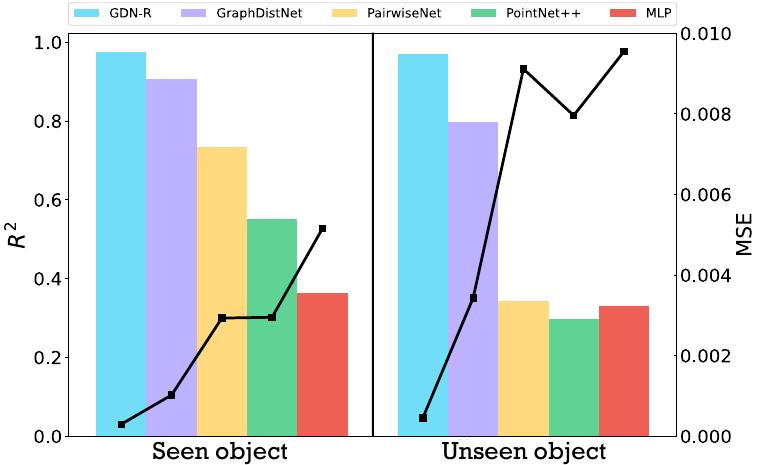}
  \caption{Comparison of GDN-R and baseline methods on object-hanging environments. We train and test each method with $26,590$ and $14,822$ random configurations of object-hook samples. For the \textit{seen category}, the training and test sets contain identical object categories, yet the shape of the object being examined in the test set is not in the training set. For the \textit{unseen category}, the training and test sets contain different category of objects. $R^2$ represents the R-squared metric that expresses the proportion of variance of a dependent variable explained by a variable in the regression model. MSE represents the mean squared error between the ground truth and the estimated minimum distance.
}
  \label{fig:bar_r2}
\end{figure}

For data collection, we downsample nodes and edges using a quadratic-error-metric mesh simplification algorithm~\cite{garland1997surface}. We then sample object-hook configurations (i.e., pose) using two strategies: 1) random pose sampling and 2) sequential sampling along randomized trajectories. For the random pose sampling, we randomly sample the position and orientation of a chosen object around a selected hook. We sample four random object poses at \SI{0.1}{\cm} intervals from the hook's surface within a \SI{10}{\cm} range in Cartesian space around the hook. Therefore, we obtain a total of $3 \times 10^5 $ pose data for $150$ object-hook pairs. For the sequential sampling, we first generate $10$ paths using the rapidly-exploring random tree (RRT) algorithm~\cite{lavalle1998rapidly}, with randomized start and goal configurations. We therefore sample object poses along the trajectory at a minimum distance of $0.02$ from the goal pose. Note that the distance metric combines both position distance $d_{p}$ and quaternion distance $d_{q}$ as follows: $0.6\cdot d_{p} + 0.4\cdot d_{q}$. We randomly scale the geometries in the samples from each strategy by a factor ranging from $1$ to $10$. We also filter out data that fall outside the range of $[0.2, 5]$\SI{}{\cm}. This process results in $41,412$ samples. We then create two benchmark datasets:
%
\begin{itemize}[leftmargin=*]
    \item \textit{Seen category}: We split object shapes using a $7:3$ ratio, irrespective of category, with $70\%$ allocated for training. 
    \item \textit{Unseen category}: We split object categories with a $3:2$ ratio for training and testing datasets, respectively. For training, we use categories including \textit{cap}, \textit{headphone}, and \textit{bag}, leaving the remaining categories for testing.
\end{itemize}
Both training and test sets consist of $26,590$ and $14,822$ samples, respectively.



We then evaluate our method with four baseline methods:
\begin{itemize}[leftmargin=*]
\item GraphDistNet~\cite{kim2022graphdistnet}: A graph-based estimator that utilizes a graph-attention mechanism to create a connected graph.
\item PairwiseNet~\cite{kim2023pairwisenet}: A point-cloud-based estimator that decomposes robot meshes and computes pairwise minimum distances between parts.
\item Pointnet++~\cite{qi2017pointnet++}: A point-cloud-based estimator encoding with Pointnet++ set-abstraction as used in \cite{you2021omnihang, murali2022cabinet}.
\item MLP: An $(16, 16)$ MLP-based distance estimator, with dropout layers, that takes vertices as inputs.
\end{itemize}
%

\subsection{Qualitative study}
We then demonstrate the applicability of our collision-distance estimation method to TO. The TO target is a cup-hanging task, where a UR5e manipulator grabs the cup and hangs the handle on a U-shaped hook. As two input graphed geometries, we use pre-defined meshes in the OmniHang benchmark. We then generate a $20$ length of collision-free 6D Cartesian path as a seed trajectory, using RRT. 

Given the initial path, We generate an optimal Cartesian trajectory via TO and calculate joint configurations using inverse Kinematics. For the optimization, we use two cost functions: \textit{constraint} and \textit{objective} costs.
The \textit{constraint} cost induces paths not to collide with obstacles by setting $constraint\_cost = w_c\sum_t\min(d_{safety} - \hat{d}_t, 0)$ where $w_c$ is a weight scale, $d_{safety}$ is the minimum safety distance, and $\hat{d}$ represents the minimum collision distance estimated from our distance model. On the other hand, the \textit{objective} cost represents the sum of distances between two subsequent points along the generated path. Combining two costs, we generate a smooth trajectory using the Adam optimizer. For the guidance of TO, we provide the collision gradient (i.e., the gradient of the collision distance with respect to the cup pose) via automatic differentiation. This is similar to the TO setup in \cite{kim2022graphdistnet, zhi2022diffco}.

\begin{table}[!t]
\small
\centering
\begin{tabular}{@{}l|c c  | c c c  c@{}}
\hline
& \multicolumn{2}{c}{Seen category} & \multicolumn{2}{c}{Unseen category}\\
\hline
Method & MSE & $R^2$   & MSE & $R^2$ \\  
\hline 
\textit{GDN-R ($\tau$=2.5)}  & \textbf{2.83e-4} 	&\textbf{0.974}	 & \textbf{3.91e-4} 	&\textbf{0.965}	 \\
\textit{GDN-R ($\tau$=5.0)}  & 3.69e-4 	&0.966		 & 5.75e-4 	&0.947	 \\
\textit{GDN-R (annealing*)}  & 2.83e-4 	&\textbf{0.974}		 & 4.09e-4 	&0.963 \\
\textit{GDN-R (no randomness)}  & 8.31e-4 	&0.923		 &9.7e-4 	&0.914	 \\
\textit{GraphDistNet}  & 1.02e-3 	&0.907		 & 2.15e-3 	&0.796	 \\
\hline
\end{tabular}
\caption{Comparison of rewiring strategies. For the \textit{unseen category}, we use the \textit{mug} and \textit{scissor} categories of objects as test objects. *-We annealed $\tau$ from $10.0$ to $0.1$.}
\label{table:model_comp}
\vspace{-5mm}
\end{table}
\section{Evaluation}

\begin{figure}[h]
 \center
  \includegraphics[width=\linewidth]{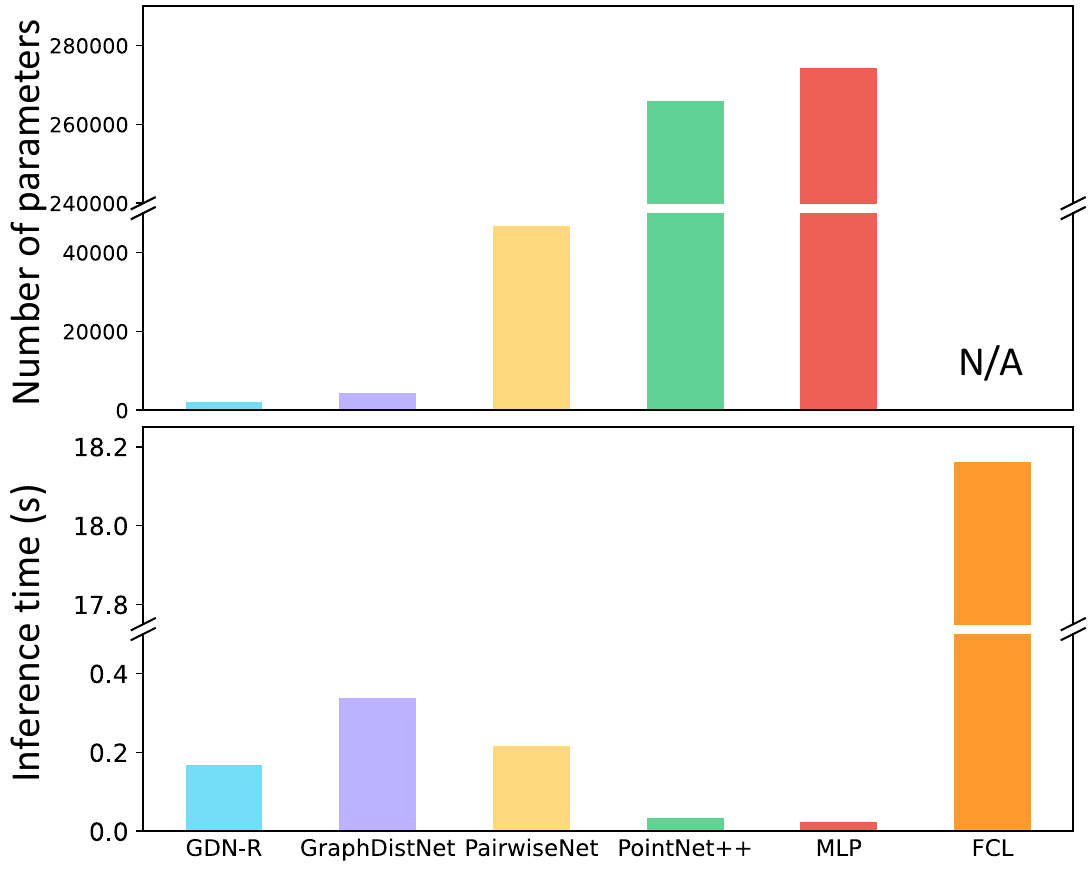}
  \caption{
    Comparison of the inference time and the number of model parameters. The average inference time is the mean time taken by each model to estimate the minimum distance, across $11$ batches of $1,024$ tasks, totaling $11,264$ tasks.
}
  \label{fig:comp_time_params}
\end{figure}

\begin{figure}[t]
 \center
  \includegraphics[width=0.75\linewidth]{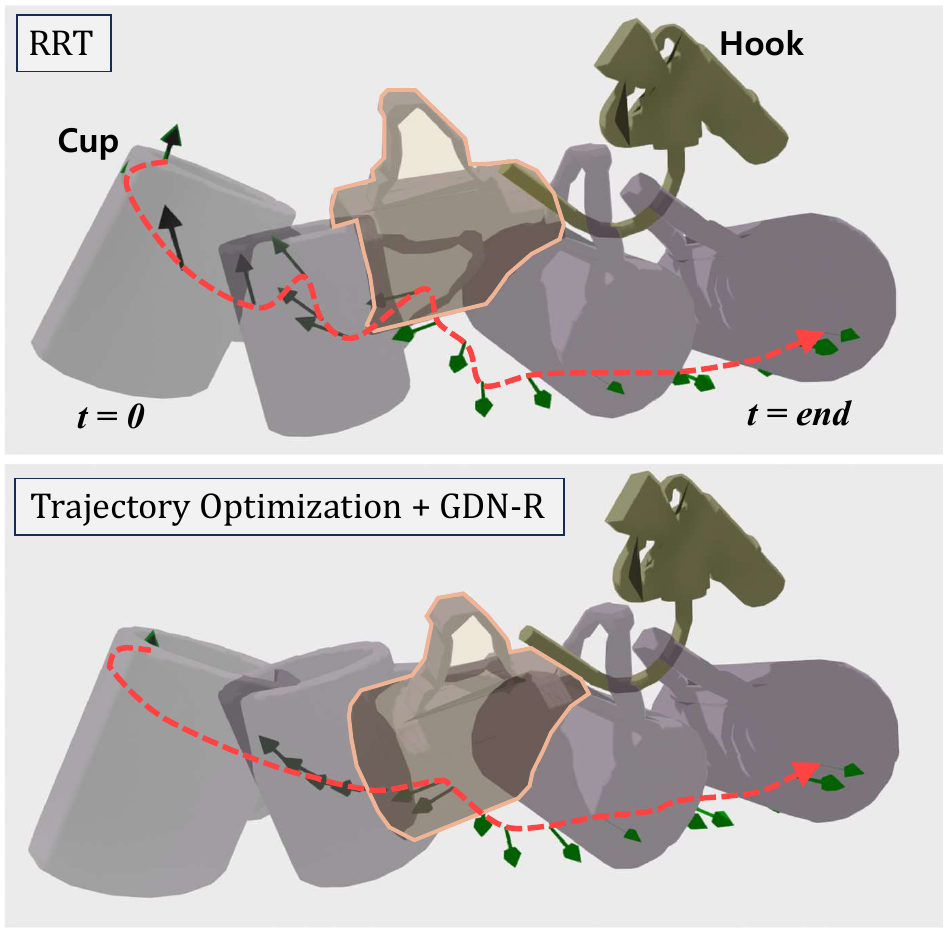}
  \caption{
    Demonstration of a simulated object-hanging task using the UR5e robot. (\textbf{Top}) The RRT-based initial trajectory is jerky, unnatural behavior. (\textbf{Bottom}) Trajectory optimization with GDN-R produces a smooth and feasible path. 
  } 
  \label{fig:demo}
\end{figure}

We first analyze the estimation accuracy of GDN-R and baselines via two benchmark tasks. Fig~\ref{fig:bar_r2} shows our method outperforms four baseline methods in both \textit{seen category} and \textit{unseen category} of benchmark tasks. Our method consistently exhibits the lowest mean squared error (MSE) as well as the highest R-squared ($R^2$) score even tested with \textit{unseen category} of objects. Note that a higher $R^2$ indicates a stronger fit for the regression model and the value is in the range of $[0,1]$. This result indicates not only the superior accuracy of GDN-R, but also its generalizability that conventional data-driven distance estimators could now show. Our previous GraphDistNet shows the second-best performance, which indicates our new rewiring algorithm effectively reconfigured the connected graphs. In addition, point-representation-based approaches resulted in significantly degraded performance than graph-based approaches. This indicates edge information plays an important role in accurate estimation. Although PairwiseNet showed adaptation performance trained with more than 1 million data in the literature, our evaluation uses a significantly smaller number of samples (i.e., $26,590$).

Table~\ref{table:model_comp} shows the ablation studies varying rewiring strategies in GDN-R. The first three strategies that utilize the Gumbel top-$k$ relaxation show superior estimation performance than the unperturbed or attention score-based rewiring methods. This indicates that probabilistic rewiring helps not only selecting the effective interconnection nodes but also expediting the exploration process using randomness. However, the randomness does not always enhance the quality of rewiring since the case $\tau=5.0$ shows performance degradation as the temperature $\tau$ increases. Overall, the probabilistic graph rewiring with small randomness enables GDN-R to find more effective interconnection nodes. 

The efficiency of the distance estimation model is also an important factor for applications. Fig~\ref{fig:comp_time_params} shows the comparison of the inference time and the number of parameters when inferring the minimum distance of $11,264$ tasks. The average inference time is the mean time taken by each model to estimate the minimum distance, across $11$ batches of $1,024$ tasks, totaling $11,264$ tasks. All of the parallelizable neural net-based methods, which allow batch computations, show significantly higher efficiency than that of the flexible collision library (FCL)~\cite{pan2012fcl}, which is the traditional single-thread search method. GDN-R records two times improved computation cost with half the amount of model parameters, compared to the previous graph-based method, GraphDistNet. Although this is slower than PointNet++ or MLP-based approaches, the batch computation speed is fast enough and the number of parameters is $10\%$ of the last two models. 

We finally demonstrate the TO with GDN-R using a UR5e robot and a cup-hanging task. Fig.~\ref{fig:demo} first shows the RRT-based initial trajectory that resulted in jerky, unnatural behavior. By optimizing the initial trajectory, our method with TO successfully produces a smooth and feasible path avoiding the contact between the cup and the hook.

\section{Conclusion}
We proposed GDN-R, a graph-based distance estimation network with a layer-wise probabilistic rewiring method. Our method constructs a message-passing effective connected graph selecting interconnection nodes via differentiable Gumbel-top-$k$ procedures. Our object-hanging benchmark evaluation shows the superior accuracy and generalizability of collision-distance estimation performance compared to state-of-the-art baseline methods, even in unforeseen 3D environments. We also demonstrated the adoption of trajectory optimization toward robotic manipulation tasks. 






\bibliographystyle{ieeetr}
\bibliography{references}

\end{document}